\documentclass{article}

\usepackage[preprint]{corl_2021} %
\usepackage{epsfig}
\usepackage{graphicx}
\usepackage{graphbox}
\usepackage{amsmath}
\usepackage{amssymb}
\usepackage{booktabs}
\usepackage{multirow}
\usepackage{url}
\usepackage[noabbrev,capitalise]{cleveref}
\usepackage[export]{adjustbox}
\usepackage[shortlabels]{enumitem}
\usepackage{caption}

\title{Rethinking Trajectory Forecasting Evaluation}

\author{
  Boris Ivanovic \hspace{0.5cm} Marco Pavone\\
  NVIDIA Research\\
  \texttt{\{bivanovic, mpavone\}@nvidia.com} \\
}

\begin{document}
\maketitle

\begin{abstract}
Forecasting the behavior of other agents is an integral part of the modern robotic autonomy stack, especially in safety-critical scenarios with human-robot interaction, such as autonomous driving. In turn, there has been a significant amount of interest and research in trajectory forecasting, resulting in a wide variety of approaches. Common to all works, however, is the use of the same few accuracy-based evaluation metrics, e.g., displacement error and log-likelihood. While these metrics are informative, they are task-agnostic and predictions that are evaluated as equal can lead to vastly different outcomes, e.g., in downstream planning and decision making. In this work, we take a step back and critically evaluate current trajectory forecasting metrics, proposing task-aware metrics as a better measure of performance in systems where prediction is being deployed. We additionally present one example of such a metric, incorporating planning-awareness within existing trajectory forecasting metrics.
\end{abstract}

\keywords{Evaluation Metrics, Trajectory Forecasting, Autonomous Vehicles}

\section{Introduction}
Predicting the future behavior of surrounding agents is a necessary capability for modern robotic systems, especially as many autonomous systems are increasingly being deployed alongside humans in domains such as autonomous driving \cite{LefevreVasquezEtAl2014,BrouwerKloedenEtAl2016}, service robotics \cite{KrusePandeyEtAl2013,ChikYeongEtAl2016,LasotaFongEtAl2017}, and surveillance \cite{MorrisTrivedi2008,MurinoCristaniEtAl2017,HirakawaYamashitaEtAl2018}.
In particular, there has been a significant interest in trajectory forecasting within the autonomous driving community, with many major organizations incorporating behavior prediction within their autonomous vehicles' software stack \cite{GMSafety2018,UberATGSafety2020,LyftSafety2020,WaymoSafety2021,ArgoSafety2021,MotionalSafety2021,ZooxSafety2021,NVIDIASafety2021}.
As a result, it is important to accurately evaluate the performance of forecasting systems prior to their use. 

\begin{figure}[t]
    \centering
    \includegraphics[width=\linewidth]{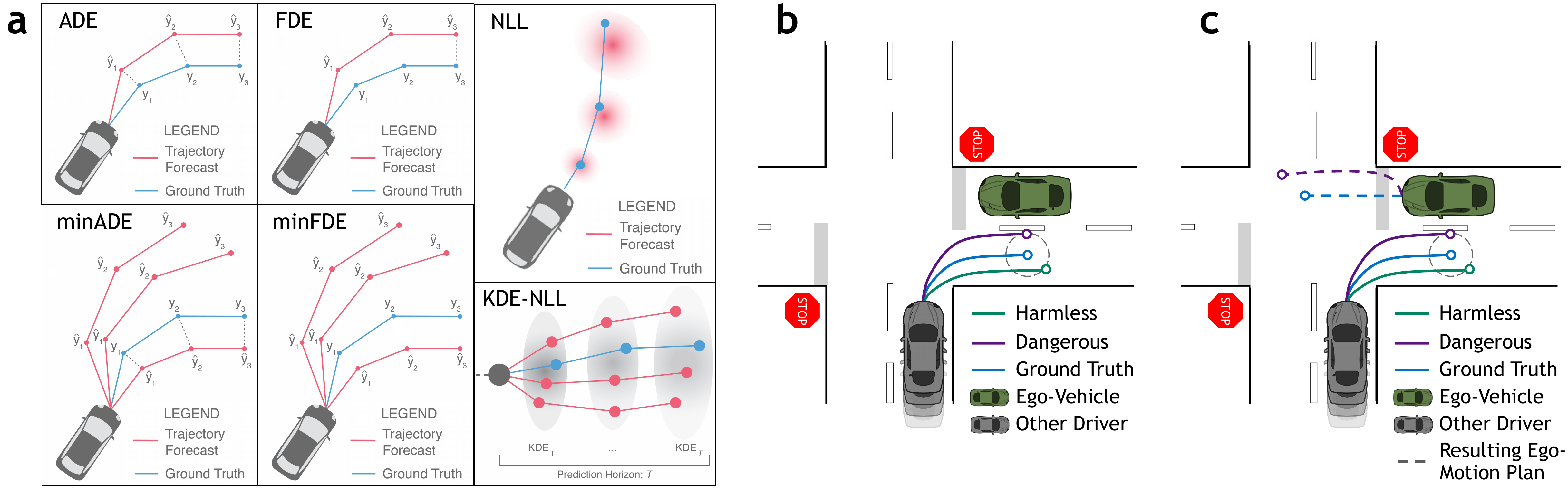}
    \caption{\textbf{(a)} 
    Accuracy-based metrics broadly evaluate how similar a forecast trajectory (in red) is to the ground truth (in blue).
    \textbf{(b)} A human driver (gray) is about to turn right next to an autonomous vehicle (green). Two predictions (solid green and purple lines) of the human are also shown, which have the same metric accuracy as they are equidistant from the ground truth future (solid blue line).
    \textbf{(c)} Although metrically equal, the purple prediction causes a safety-preserving maneuver (dashed purple) while the other does not affect the autonomous vehicle's motion plan (dashed blue).}
    \label{fig:hero}
    
    \vspace*{-0.5cm}
    
\end{figure}

To date, nearly all works have relied on accuracy-based metrics such as average or final displacement error (ADE/FDE), negative log-likelihood (NLL), and other geometric or probabilistic quantities (see Table 1 of \cite{RudenkoPalmieriEtAl2019} for a comprehensive list, and \cref{fig:hero} (a) for illustrated examples). At their core, accuracy-based metrics compare a model's predicted trajectory (or distribution thereof) with the ground truth future trajectory realized by an agent,
producing a value that quantifies how similar the two are. Comparing trajectories solely based on accuracy, however, does not consider downstream ramifications, and errant predictions with equal metric inaccuracy can lead to vastly different outcomes, an example of which is illustrated in \cref{fig:hero} (b) and (c).

{\bf Contributions.}
Towards this end, our contributions are twofold. First, we 
argue for the use of task-aware metrics to evaluate methods in a manner that better matches the systems in which they are deployed. Second, we present a novel planning-aware prediction metric as an example of a task-aware metric for prediction methods whose outputs are used to inform downstream planning and decision making, an arrangement commonly found in modern robotic autonomy stacks (e.g., \cite{WaymoSafety2021}).

\section{Related Work}\label{sec:litreview}

{\bf Trajectory Forecasting Evaluation.} There has been a significant surge of interest in trajectory forecasting within the past decade, spawning a diverse set of approaches combining tools from physics, planning, and pattern recognition \cite{RudenkoPalmieriEtAl2019}. Accordingly, there have been many associated thrusts in developing prediction metrics that accurately evaluate these methods \cite{MorrisTrivedi2008,ZhangHuangEtAl2006,Zheng2015,QuehlHuEtAl2017}. Overall, two high-level classes of metrics have emerged: geometric and probabilistic. Geometric metrics (e.g.,~ADE and FDE) compare a single predicted trajectory to the ground truth, whereas probabilistic metrics (e.g., minimum ADE/FDE, NLL, kernel density estimate (KDE)-based NLL \cite{IvanovicPavone2019}) compare predicted distributions or sets of trajectories to the ground truth, taking into account additional information such as 
variance. A few examples of existing metrics are depicted in \cref{fig:hero} (a).

While existing metrics are useful for evaluating the performance of trajectory forecasting methods in isolation, there are important 
considerations that arise during real-world deployment. Some examples include handling perception uncertainty in prediction \cite{LuoYangEtAl2018,DiHarakehEtAl2020,IvanovicLeeEtAl2021} and integrating prediction and planning \cite{KrusePandeyEtAl2013,SchmerlingLeungEtAl2018,NishimuraIvanovicEtAl2020,IvanovicElhafsiEtAl2020,SchaeferLeungEtAl2021}).
Most importantly, in this work we focus on the fact that
\emph{prediction errors are asymmetric in the real world}, i.e., predictions with the same metric accuracy may lead to vastly different outcomes,
an example of which is illustrated in \cref{fig:hero} (b) and (c). Accounting for such asymmetry is of critical importance now that prediction algorithms are executed during planning in a closed-loop fashion (e.g., \cite{SchaeferLeungEtAl2021,TolstayaMahjourianEtAl2021}) and
deployed in real-world, safety-critical settings.

{\bf Task-Aware Metrics.} Comparatively, there are much fewer works on task-aware metrics for components of the autonomy stack (e.g., detection, tracking, prediction, planning). One notable example
is the Planning KL Divergence (PKL) metric \cite{PhilionKarEtAl2020}. PKL measures how similar an object detector's outputs are to the ground truth by computing the difference between an ego-vehicle's plan using predicted detections and its plan using ground truth detections.
A downside of this strategy is that it relies on the performance of a specific neural planner, which makes it difficult to determine if regressions in performance are caused by the detector or planner (e.g., could a better planner handle a wider range of detection errors? Do certain detectors only work well with certain planners?). Further, 
tools such as trajectory optimization, graph search, variational methods, and sampling-based motion planning are significantly more common than neural planning in real world systems \cite{PadenCapEtAl2016}.
In a typical modern autonomous driving system \cite{GMSafety2018,UberATGSafety2020,LyftSafety2020,WaymoSafety2021,ArgoSafety2021,MotionalSafety2021,ZooxSafety2021,NVIDIASafety2021}, a learned prediction model forecasts the future trajectories of agents around an ego-vehicle and a downstream planner uses the predictions to plan a smooth, collision-avoiding path to a goal state.
An intuitive initial idea for evaluating the predictor in this system is to compare the outputs of the planner when using the predicted trajectories versus the ground truth future trajectories, similar to the PKL metric~\cite{PhilionKarEtAl2020}. We will discuss the shortcomings of this strategy in the following section.

\section{Task-Aware Prediction Metrics}\label{sec:method}
In this section, we advocate for a set of core considerations that a task-aware prediction metric should address and apply them to develop a proof-of-concept planning-aware prediction evaluation framework.

{\bf Core Desiderata.} %
In particular, we advocate that task-aware trajectory forecasting metrics be:
\begin{enumerate}[(1),itemsep=0.05em,topsep=0pt]
    \item Able to capture asymmetries that arise in downstream tasks.
    \item Task-aware and method-agnostic.
    \item Computationally feasible to compute.
    \item Interpretable.
\end{enumerate}
The first consideration directly addresses the core shortcoming of existing metrics, and is the main motivation for this work. In particular, it would be ideal to explicitly distinguish between harmless and catastrophic prediction errors, penalizing them in a similarly-asymmetric manner (i.e., weighing worse prediction outcomes more heavily).
The second desideratum is more nuanced, and requires task-aware metrics to be decoupled from any one particular method for that task. For example, a planning-aware prediction metric should not rely on a specific planner (e.g., FMT*~\cite{JansonSchmerlingEtAl2015}, RRT*~\cite{KaramanFrazzoli2011}) in its computation. The reason for this is that it harms applicability (not all practitioners use the same planner in their system of interest) and introduces biases (a specific planner may require a specific output format from an upstream prediction algorithm, which unfairly advantages methods that produce the same output format).
Importantly, these shortcomings directly apply to the intuitive predictor-planner evaluation idea described in \cref{sec:litreview}, making it arguably unsuitable for real-world use.
The third consideration ensures that practitioners are able to obtain metric values efficiently over modern large-scale datasets.
Finally, interpretability is important for any evaluation metric \cite{Doshi-VelezKim2017}, allowing users to understand and contextualize the performance of their method.

\subsection{Planning-Aware Prediction Evaluation as a Proof of Concept}
We present a proof-of-concept instantiation of a task-aware prediction metric that fulfills the aforementioned desiderata. In particular, we develop a planning-aware prediction metric for autonomous driving that evaluates trajectory forecasting models based on their effect on an ego-vehicle's downstream motion planning. Specifically, our method identifies and heavily weighs prediction errors that would be catastrophic. At a high-level, our approach leverages human trajectory datasets (e.g., \cite{CaesarBankitiEtAl2019,HoustonZuidhofEtAl2020,SunKretzschmarEtAl2020}) to learn a planning cost function whose sensitivities to prediction outputs determine which agents most influence planning. These sensitivities can then be used to inject task-awareness within existing metrics (e.g., by weighing prediction accuracies based on their planning influence). We also experimentally demonstrate its performance in a simple, but illustrative, collision-avoidance scenario.

{\bf Planning-Aware Prediction Evaluation.}
Since using a specific planner would go against the second desideratum, we instead work with planning \emph{cost functions}. Cost functions are general (defining the high-level goal for a planning problem) and planner-agnostic (the same cost function can be minimized with different planners, leading to different output paths). Which cost function to use is an important consideration, and we maintain method-agnosticism by 
learning a proxy cost function whose minimization well-reproduces ego-vehicle trajectories in an autonomous driving dataset. In particular, given the $D_S$-dimensional dynamic states $\mathbf{s}^{(t)} \in \mathbb{R}^{(|\mathcal{A}|+1) D_S}$ (e.g., position, velocity, acceleration) of an ego-vehicle and other agents $a \in \mathcal{A}$ in the scene, the $D_U$-dimensional actions enacted by the ego-vehicle $\mathbf{u}_\text{R}^{(t)} \in \mathbb{R}^{D_U}$, as well as $D_P$-dimensional predictions of other agents' future positions for the next $T$ timesteps $\mathbf{\hat{s}}^{(t:T)} \in \mathbb{R}^{T \times |\mathcal{A}| D_P}$, we specify a linear cost function $c$ of the form
$c(\mathbf{s}^{(t)}, \mathbf{u}_\text{R}^{(t)}, \mathbf{\hat{s}}^{(t:T)}) = \theta^T \phi(\mathbf{s}^{(t)}, \mathbf{u}_\text{R}^{(t)}, \mathbf{\hat{s}}^{(t:T)}),$ 
where $\theta_i \in \mathbb{R}$ is the weight of the $i^\text{th}$ feature $\phi_i: \mathbb{R}^{(|\mathcal{A}|+1) D_S \times D_U} \rightarrow \mathbb{R}$. 
Our motivation for using a linear function is that it additively combines features that are considered by the ego-vehicle during planning (leading to easier interpretation \cite{Molnar2019LinReg}), and, although simple, we find that it is already useful for task-aware trajectory forecasting evaluation. We then leverage Continuous Inverse Optimal Control (CIOC) to learn the weights $\theta_i$. Importantly, since CIOC does not require globally-optimal expert trajectories in general~\cite{LevineKoltun2012}, we only assume that human trajectories are \emph{locally} optimal. 
With the learned cost function in hand, we compute its sensitivity with respect to the predicted agent locations by obtaining the 
gradient $\nabla_{\mathbf{\hat{s}}^{(t:T)}} c$. The gradient magnitudes are a measure of how sensitive the ego-vehicle's plan is to each agent's prediction, with higher values indicating more sensitivity. Finally, these values can be used to weigh predictions
within existing accuracy-based metrics. For example, planning-informed (PI) versions of accuracy-based metrics can be implemented
as:
\begin{equation}\label{eqn:pimetric}
    \text{PI-Metric} = \frac{1}{|\mathcal{A}|} \sum_{a \in \mathcal{A}} f(a, |\nabla_{\mathbf{\hat{s}}^{(t:T)}} c|) \cdot  \text{Metric}(\mathbf{\hat{s}}^{(t:T)}_a, \mathbf{s}^{(t:T)}_a),
\end{equation}
where ``Metric" is a placeholder for existing metrics (e.g., those in \cref{fig:hero} (a)) and $f$ allows for the implementation of different weighting schemes, e.g., normalization (where $f(a, g) = 1 + g_a / \sum_{a' \in \mathcal{A}} g_{a'}$) or softmax (where $f(a, g) = 1 + \exp(g_a) / \sum_{a' \in \mathcal{A}} \exp(g_{a'})$).

\begin{figure}[t]
    \centering
    \includegraphics[width=\linewidth]{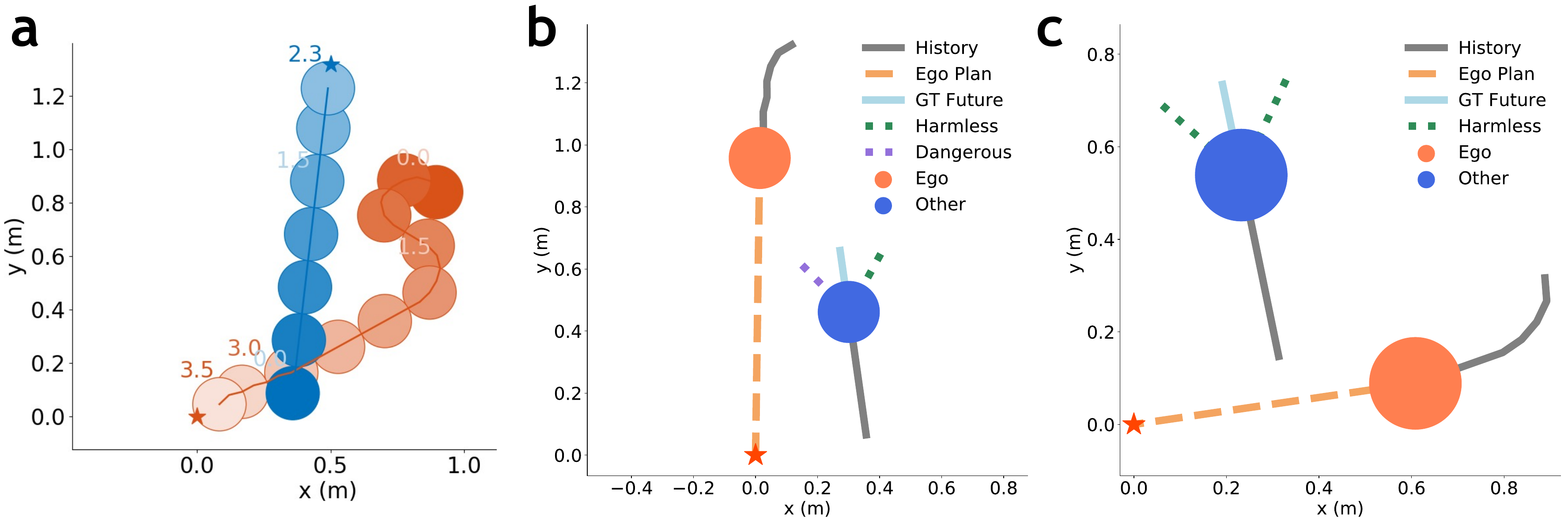}
    \caption{
    \textbf{(a)}. An ego-vehicle (orange) maneuvers to the origin while avoiding other agents (blue), lighter colors occur later. 
    Importantly, our method is able to distinguish between metrically-equal errant predictions in a planning-aware manner. In a head-on scenario \textbf{(b)}, planning sensitivities are much higher for predictions that veer into the ego-vehicle's path (purple dashed) compared to those that steer away (green dashed). Further, when it is unlikely that an agent would influence the ego-vehicle's plan \textbf{(c)}, our method yields small planning sensitivities for all predictions.}
    \label{fig:expt}
    
    \vspace*{-0.5cm}
    
\end{figure}

{\bf Illustrative Collision-Avoidance Scenario.}
We experimentally demonstrate the performance of our proposed metric augmentation in an environment where an ego-vehicle with unicycle dynamics~\cite{LaValle2006Unicycle} is tasked with reaching the origin from a random starting state while avoiding collisions with surrounding vehicles. %

Our cost function contains four terms: A goal term (squared distance between the ego-vehicle and the origin), a control term (squared magnitude of the control effort), and two collision-avoidance terms (radial basis functions centered at the other agents' current and one-step predicted positions). Then, 64 rollouts of a pre-trained 
GA3C-CADRL \cite{EverettChenEtAl2021} policy were collected in an associated collision avoidance Gym environment \cite{EverettChenEtAl2021} (\cref{fig:expt}~(a) shows one such rollout).
The cost function term weights, $\theta$, are obtained from the rollouts with CIOC~\cite{LevineKoltun2012}.
Finally, the ego-vehicle's planning sensitivity to agent predictions is obtained
with standard autodifferentiation tools~\cite{PaszkeGrossEtAl2017}. 

We analyze our method's performance in two scenarios. \cref{fig:expt} (b) shows a scene where the ego-vehicle must avoid a head-on collision with another agent to reach its goal. Two predictions with the exact same raw displacement errors (ADE = $0.075$, FDE = $0.15$) are also depicted (dashed purple and green). The resulting gradient magnitudes are $0.90$ (purple) and $0.21$ (green), sensibly indicating that the purple prediction affects the ego-vehicle's planning the most (it veers directly into the ego-vehicle's desired path). Applying \cref{eqn:pimetric} with these values yields a $25\%$ higher piADE and piFDE for the purple prediction than the green prediction, with $f(a, g) = 1 + \max(0, g_a - g_{a_{GT}})$ where $g_{a}$ and $g_{a_{GT}} = 0.57$ are the planning sensitivities of the prediction and ground truth future. Finally, in \cref{fig:expt} (c) we sanity check the performance of our method when there is little planning influence from another agent. As expected, the planning sensitivities are all near zero, yielding piADE and piFDE that match their task-agnostic counterparts ADE and FDE.

These results are encouraging as they demonstrate that our method can identify prediction outcome asymmetries and
incorporate such information within existing metrics, 
providing a measure of downstream task performance. Such information can even be used at training time as our metric is fully-differentiable, and an interesting area of future research is determining how downstream tasks influence prediction network training.

\section{Conclusion and Future Directions}

In this work, we advocate for the incorporation of task-awareness in trajectory forecasting evaluation. In particular, we show by way of example that existing metrics neglect the asymmetries of real-world prediction outcomes, outline four core considerations that any task-aware metric should address, and provide a proof-of-concept framework and instantiation illustrating one way that task-awareness could be injected into existing metrics. Looking forward, this work further serves as a call to action for additional research in task-aware prediction metrics, especially now that the field has reached a level of maturity where state-of-the-art methods are commonly deployed in real-world, safety-critical settings. Evaluation methods should reflect this maturity and deployment-readiness.

Looking beyond prediction, enabling task-aware evaluation for other autonomy stack components (e.g., detection and tracking) is another strong future direction as it would enable for the parallel co-design of autonomy modules with increased confidence in their integrated performance (since all components are already evaluated against each other during development).

\acknowledgments{We thank Andrea Bajcsy for her manuscript feedback and revisions, as well as Karen Leung, Edward Schmerling, Shie Mannor, Nikolai Smolyanskiy and the rest of the NVIDIA AV Prediction team for many important discussions that shaped this work along the way.}

\bibliography{ASL_papers,main}

\end{document}